\RequirePackage{fix-cm}

\documentclass[reqno]{amsart}
\usepackage{amsaddr}
\usepackage{graphicx}
\usepackage{url}
\usepackage{ulem}
\usepackage[nocompress]{cite}
\usepackage[utf8]{inputenc}
\usepackage{amssymb}
\usepackage{multirow}
\usepackage[linesnumbered,ruled]{algorithm2e}
\usepackage{tcolorbox}
\usepackage{setspace}
\usepackage{amsmath}
\usepackage{amsfonts}
\usepackage{graphics}
\usepackage{enumitem}
\usepackage{xcolor}
\usepackage{hyperref}
\usepackage{multirow}
\usepackage{booktabs}
\usepackage{adjustbox}
\usepackage{pgfplots}
\usepackage{subcaption}

\newcommand{\R}{{\mathbb{R}}}

\newcommand{\bfc}{{\bf c}}

\newcommand{\bfx}{ {\bf x}}
\newcommand{\bfy}{ {\bf y}}

\newcommand{\bfD}{{\bf D}}

\newcommand{\bfG}{{\bf G}}

\newcommand{\bfI}{{\bf I}}
\newcommand{\bfJ}{{\bf J}}
\newcommand{\bfK}{{\bf K}}

\newcommand{\bfS}{{\bf S}}

\newcommand{\bfW}{{\bf W}}

\newcommand{\bftheta}{{\boldsymbol{\theta}}}
\newcommand{\bfOmega}{{\boldsymbol{\mathrm \Omega}}}

\usepgfplotslibrary{fillbetween}
\pgfplotsset{compat=1.17}

\makeatletter
\def\paragraph{\@startsection{paragraph}{4}%
  \z@\z@{-\fontdimen2\font}%
  {\normalfont\bfseries}}
\makeatother

\begin{document}
\title[Quantized CNNs through the Lens of PDEs]{Quantized Convolutional Neural Networks through the Lens of Partial Differential Equations}
\author{Ido Ben-Yair, Gil Ben Shalom, Moshe Eliasof \and \\Eran Treister$^1$}
\address{Ben-Gurion University of the Negev}
\email{$^1$erant@cs.bgu.ac.il}
\thanks{The research reported in this paper was supported by the Israel Innovation Authority through the Avatar
consortium, and by grant no. 2018209 from the United States - Israel
Binational Science Foundation (BSF), Jerusalem, Israel. ME is supported by Kreitman High-Tech scholarship. The authors thank the Lynn and William Frankel Center for Computer Science at BGU}

\maketitle

\begin{abstract}
Quantization of Convolutional Neural Networks (CNNs) is a common approach to ease the computational burden involved in the deployment of CNNs, especially on low-resource edge devices. However, fixed-point arithmetic is not natural to the type of computations involved in neural networks.  
In this work, we explore ways to improve quantized CNNs using PDE-based perspective and analysis. 
First, we harness the Total Variation (TV) approach to apply edge-aware smoothing to the feature maps throughout the network. This aims to reduce outliers in the distribution of values and promote piece-wise constant maps, which are more suitable for quantization. 
Secondly, we consider symmetric and stable variants of common CNNs for image classification, and Graph Convolutional Networks (GCNs) for graph node-classification. We demonstrate through several experiments that the property of forward stability preserves the action of a network under different quantization rates. As a result, stable quantized networks behave similarly to their non-quantized counterparts even though they rely on fewer parameters. We also find that at times, stability even aids in improving accuracy. These properties are of particular interest for sensitive, resource-constrained, low-power or real-time applications like autonomous driving.
\end{abstract}

\section{Introduction}
\label{sec:intro}
Deep neural networks have demonstrated superiority in solving many real world problems. In particular, Convolutional Neural Networks (CNNs)~\cite{LeCun1990} are among the most effective approaches for processing structured high-dimensional data for a wide array of tasks, from speech recognition~\cite{RainaEtAl2009} to image classification ~\cite{KrizhevskySutskeverHinton2012} and segmentation~\cite{ronneberger2015}. 
Consequently, the demand for the deployment of CNNs on resource-constrained devices is constantly increasing (e.g., at the edge as part of larger systems and services), where at the same time, CNNs grow larger than ever~\cite{brown2020language}, to improve upon state-of-the-art models.

Despite their enormous success, CNNs still face critical challenges. First, it is well-established that the predictions obtained by deep neural networks can be highly sensitive to perturbations of the input~\cite{MoosaviDezfooli:2015wj,goodfellow2018making}. Further, to reach high accuracy figures, CNNs often require extremely deep and wide architectures, in terms of the number of hidden layers and channels, respectively. This leads to a tremendous number of parameters, imposing a high computational cost. This makes the deployment of CNNs on resource-constrained devices prohibitive, especially for mission-critical applications such as autonomous driving.

In recent years, scholars have sought to alleviate the computational burden of CNNs by developing specialized hardware to support these computational demands~\cite{hardware_survey} as well as model compression methods in order to reduce them~\cite{compression_survey}. It is common to distinguish between different approaches such as pruning methods~\cite{pruning_survey}, knowledge distillation~\cite{distillation_survey}, neural architecture search (NAS)~\cite{nas_survey} and network quantization~\cite{hubara2017quantized}, which is the focus of this work. Quantization methods enable neural networks to carry out computations with fixed-point operations rather than floating-point arithmetic. This contributes to the efficiency of the networks and reduces their memory footprint. However, this often comes at the cost of accuracy~\cite{compression_survey}. This trade-off guided the proposal of various quantization schemes designed for different use-cases. 

Concurrently, several authors have recently established a direct link between CNNs and partial differential equations (PDEs)~\cite{E2017,ChaudhariEtAl2017,HaberRuthotto2017}. The connection is two-fold. First, CNNs filter input features with multiple layers, employing both elementwise non-linearities and affine transformations. These transformations are based on convolution operators with compactly supported filters~\cite{goodfellow2016wc}, which can be seen as linear combinations of the finite difference discretizations of spatial derivative operators~\cite{ruthotto2019deep,ephrath2020leanconvnets,eliasof2020diffgcn}. Furthermore, the popular residual network architectures (ResNets)~\cite{he2016deep} can be interpreted as a discrete time integration of a non-linear ODE using the forward Euler scheme. This conceptual framework gives rise to many interesting questions, one of which is: What is the importance of the forward stability of CNNs, and can we create analogues to stable time integration through Courant-Friedrichs-Lewy conditions, to make CNNs more robust to errors? Specifically, here we consider round-off and quantization errors under these analogues. 

In this work we wish to examine the concept of quantized CNNs using tools from the theoretical analysis of PDEs. To this end, we begin by considering that forward time integration in PDEs inevitably generates error at every step (whether round-off or discretization error). To satisfy forward stability, this error must decay from time step to time step, such that the integration error is bounded, resulting in a discrete approximation close to the real continuous solution of the PDE. A similar argument can be made for real-valued CNNs vs their quantized counterparts.

\paragraph{Our Contribution}
A significant contribution of this work is the promotion of stability in quantized CNNs via small changes to commonly known architectures.
Our first proposal considers the Total Variation (TV) method~\cite{RudinOsherFatemi1992}, commonly used for image denoising via edge-preserving anisotropic diffusion. Essentially, a TV diffusion process is known to promote smooth piecewise-constant images, which is natural for quantization. We show that by integrating TV smoothing steps into existing network architectures (as a non-pointwise activation), we improve the performance of CNNs in classification and semantic segmentation tasks.
Next, we examine the behaviour of quantization under symmetric and stable, heat equation-like CNNs~\cite{HaberRuthotto2017,alt2021translating,alt2021connections}. We show that the quantization process produces significantly lighter-weight networks, in terms of storage and computation, while only incurring a minimal loss of accuracy. We develop the appropriate architectural variants for ResNet-like CNNs as well as GCNs such as~\cite{eliasof2021pdegcn}.
Finally, we study the consistency of quantization methods in CNNs. That is, given two networks of identical architecture, where one is quantized and the other is real-valued, it is desirable that their feed-forward computations proceed in a similar manner in terms of their hidden feature maps, notwithstanding quantization error. If the latter condition is satisfied, then the two networks are considered to be \textit{consistent}. This consistency can be a key attribute in the construction of quantized variants of CNNs for mission-critical tasks such as autonomous driving. 
To this end, we measure the aforementioned similarity using symmetric (potentially stable) and standard (potentially non-stable) networks. Our experiments indicate that symmetric and stable networks achieve better consistency.

To summarize, in this work we harness PDE-inspired stability theory to obtain reduced error in activation maps, where this error results from quantization of the parameters and activation maps of the neural network models. We show that \emph{quantized networks are expected to perform better and more in line with their full-precision counterparts when stable architectures are used, rather than unstable architectures}. In our opinion, such quantized neural networks are more trustworthy for sensitive tasks deployed on edge devices.

\paragraph{Organization of this paper}
This paper is organized as follows: in section \ref{sec:related_work} we discuss related work, and in section \ref{sec:background} we present relevant concepts and definitions that form the background of our work. In section \ref{sec:tv_cnn}, we present our first proposal of using a Total Variation smoothing layer as an additional activation function. This is followed by our modification of neural network architectures to gain forward stability in networks based on ResNet, MobileNetV2 and PDE-GCN~\cite{eliasof2021pdegcn}, presented in section \ref{sec:stable_nets}. Finally, we present numerical experiments to support our theoretical discussions in section \ref{sec:experiments}.

\section{Related Work}
\label{sec:related_work}

\paragraph{Neural Network Quantization}
Two methods are discussed frequently: post-training quantization and quantization-aware training. In post-training quantization, we decouple the training of the model and the quantization of its weights and activations. That is, quantization is performed after training is completed. This approach is most useful when the training data is no longer available after the training ~\cite{soudry1,postq1,nagel2019dfq}. Alternatively, quantization-aware training performs the training and quantization simultaneously~\cite{hubara2017quantized}. This approach requires access to the training data but typically yields better performance.

Quantization schemes can be classified as uniform or non-uniform. Uniform quantization divides the real domain into equally sized bins with simple arithmetic, whereas non-uniform methods admit bins of varying size at the cost of more expensive computation. Uniform quantization is favored for utilizing hardware more efficiently~\cite{han2016deep,yin2019blended}, and such quantization schemes were extensively developed in~\cite{drfn,zhang2018lqnets,choi2018pact,jin2020scaleadjusted}, learning per-layer quantization parameters such as scaling.
Non-uniform quantization is superior for performance-oriented applications~\cite{li2019apot,jung2019learning}. Furthermore, the bit allocation (number of bits used to represent values) can remain constant for all layers or vary throughout the network~\cite{bodner2021gradfreebits}. In this work we consider a per-layer, uniform and quantization-aware training method, using the same bit allocation for the whole network.

One problem of interest in this paper is semantic segmentation~\cite{chen2017atrous}, for which we consider the use of quantization. In the works~\cite{uhlich2020mpd, xu2018biomedical} only the weights are quantized for medical image segmentation, as an attempt to remove noise rather than to gain computational efficiency. The work~\cite{yuang2021zaq} shows a post-training quantization, including fine-tuned semantic segmentation results.
\bigskip
\paragraph{CNNs as Discretized PDEs}
As mentioned before, the link between neural networks and PDEs has been established previously by several authors. E~\cite{E2017} suggests a profound connection between deep learning architectures and dynamical systems. The work~\cite{ChaudhariEtAl2017} studies the training process of Stochastic Gradient Descent from the perspective of stochastic PDEs. In the work~\cite{ruthotto2019deep} the authors derive an entire family of PDE- and ODE-motivated neural network architectures. In~\cite{HaberRuthotto2017}, neural network architectures are designed based on non-linear ODEs to prevent vanishing and exploding gradients. 
In~\cite{zhang2020forward} the authors develop a connection between residual networks and optimal control problems and analyze the stability of the network with these problems in mind. In~\cite{gunther2020layer} the authors develop a layer-parallel approach to train ResNets based on parallel-in-time PDE solvers. Taking a different viewpoint on ResNet architectures,~\cite{haber2019imexnet} derive a network architecture using an implicit-explicit formulation to address stability and field-of-view issues standard CNNs. Ephrath et al.~\cite{ephrath2020leanconvnets} present a set of sparse and lean convolution operators, motivated by finite difference discrete operators, to reduce running times in inference and training. Finally, Total Variation was explored in the context of deep neural networks, such as an approach suggested very recently by~\cite{alt2021connections}. All these methods did not consider the framework of quantized neural networks. 
\bigskip
\paragraph{Neural Network as Continuous Functions}
Another possible approach is to employ a time integration scheme by viewing the network as a continuous function. For example,~\cite{chen2018neural} view the forward propagation of a neural network, at the limit case of having infinitely many layers, as analogous to an ODE, and implement inference by calling a forward Euler ODE solver. They then derive a tractable procedure to compute the gradient w.r.t the loss as a solution of the adjoint equation by another call to an ODE solver. Gholami et al.~\cite{Gholami2019ANODEUA} suggest a discretize-then-optimize approach in addition to reducing memory requirements by checkpointing. In this work, we consider the discretize-then-optimize approach and use the ResNet architecture as a discretization of a continuous PDE for the purpose of applying theory of PDEs in the case of quantization.
\bigskip
\paragraph{Graph Convolutional Networks}
Graph Convolutional Networks~\cite{kipf2016semi,chen2020simple} have received similar treatment as CNNs. Authors have considered the connection between GCNs and PDEs on more spatially complex domains. Eliasof et al.~\cite{eliasof2020diffgcn} consider graph convolutions based on discretizations of the gradient and divergence differential operators for arbitrary graphs. Xhonneux et al.~\cite{xhonneux2020contgnn} consider GCN models motivated by various ODEs such as ones modelling the spread of epidemics, while Chamberlain et al.~\cite{chamberlain2021grand} construct their architecture from diffusion equations and employ a self-attention mechanism. Thorpe et al.~\cite{thorpe2022grandplusplus} augment \cite{chamberlain2021grand} with an additional source term for classification problems with sparse labelling.
Eliasof et al.~\cite{eliasof2021pdegcn} consider both the diffusion and wave equations as bases for GCN architectures for various problems and datasets. In this paper, we follow~\cite{eliasof2021pdegcn} and demonstrate our theory using the PDE-GCN architecture.

\section{Background}
\label{sec:background}
In this section we provide a mathematical background on the two main ideas that we consider in the paper: continuous CNNs and their stable variants, as well as quantized CNNs. Following that, we provide the motivation for examining quantized CNNs through the lens of PDEs.  

\subsection{\textbf{Continuous Neural Networks and Stable ResNet Architectures}} 
\label{sub:background_continuous}
The general goal of supervised machine learning is to model a function $f : \R^n \times \R^p \to \R^m$ and train its parameters $\bftheta\in \R^p$ such that
\begin{equation}
\label{eq:interp}
    f(\bfy, \bftheta) \approx \bfc
\end{equation}
for input-output pairs $\{(\bfy^i,\bfc^i)\}_{i=1}^s$ from a certain data set $\mathcal{Y} \times \mathcal{C}$. Typical tasks include regression or classification, and the function $f$ can be viewed as an approximation of the true function that we want the machine to learn.

In learning tasks involving images and videos, the learnt function $f$ commonly includes a CNN that filters the input data.
Among the many architectures we focus on feed-forward networks of a common type called by ResNets~\cite{he2016deep}. In the simplest case, the forward propagation of a given example $\bfy$ through an $N$-layer ResNet can be written as
\begin{equation}
\label{eq:resnetintro}
    \bfx_{j+1} = \bfx_j + h F(\bfx_j, \bftheta_{j}), \quad j=0,\ldots,N-1, \quad \bfx_0 = F_{\mathrm{open}}(\bfy,\bftheta_{\mathrm{open}}).
\end{equation}
The layer function $F$ consists of spatial convolution operators parameterized by the weights $\bftheta_1, \ldots, \bftheta_N$, and non-linear element-wise activation functions. The parameter $h>0$ in this notation serves as a time step, arising from a discretization of a continuous network, as we show in the next section.
The classical ResNet layer reads 
\begin{equation}
\label{eq:resnet}
    F_{\rm ResNet}(\bfx,\bftheta) = \bfK_1 \sigma \left( \bfK_2 \bfx\right).
\end{equation}
where $\bfK_1,\bfK_2$ are two different convolution operators parameterized by $\theta$, and $\sigma$ is a non-linear activation function, like $\sigma(x) = \max\{x,0\}$, known as the Rectified Linear Unit (ReLU) function~\cite{nair2010relu}. The term $F_{\mathrm{open}}$ denotes an opening layer that typically outputs a $n_c$-channel image $\bfx_0$, where $n_c$ is greater than the number of input channels, which is typically 3 for RGB images, while possibly also reducing the size of the image. Usually, the opening layer reads  
\begin{equation}
\bfx_0 = \sigma(\bfK_{\mathrm{open}}\bfy)
\end{equation}
where $\bfK_{\mathrm{open}}$ is a convolution operator parameterized by $\bftheta_{\mathrm{open}}$ that widens the number of channels from this point onward in the network. 

Continuous neural networks have recently been suggested as an abstract generalization of the more common discrete network, where the network is viewed as a discretized instance of a continuous ODE or PDE. As shown by~\cite{ChaudhariEtAl2017,E2017,HaberRuthotto2017,chen2018neural}, Eq. \eqref{eq:resnetintro} (or a ResNet) is essentially a forward Euler discretization of a continuous non-linear ODE
\begin{equation}
\label{eq:ResNNcont}
    \partial_t \bfx(t) = F(\bfx(t), \bftheta(t)), \quad t \in [0,T], \quad \bfx(0) = \bfx_0 = F_{\mathrm{open}}(\bfy,\bftheta_{\mathrm{open}}),
\end{equation}
where  $[0,T]$ is an artificial time interval related to the depth of the network. Relating spatial convolution filters with differential operators, the authors of~\cite{ruthotto2019deep} propose a layer function representation $F$ that renders Eq. \eqref{eq:ResNNcont} similar to a parabolic diffusion-like PDE
\begin{equation}
\label{eq:doubleSym}
    F_{\rm sym}(\bfx,\bftheta) = - \bfK^\top \sigma \left(\bfK \bfx\right).
\end{equation}
For example, when $\bfK$ represents a discrete gradient operator and $\sigma(x)=x$, we obtain the heat equation under this treatment. The approach of Eq. \eqref{eq:doubleSym} is natural, as similar developments have led to several breakthroughs in image processing~\cite{weickert1998anisotropic}, including optical flow models for motion estimation~\cite{HornSchunck1981}, non-linear anisotropic diffusion models for image denoising~\cite{PeronaMalik1990,RudinOsherFatemi1992}, and variational methods for image segmentation~\cite{AmbrosioTortorelli1990,ChanVese1999}. 
To best balance the network's representational abilities with its computational cost, a ``bottleneck'' structure is commonly used where $\bfK$ has a different number of input and output channels.

Generally speaking, the training of the neural network model consists of finding parameters $\bftheta$ such that Eq.~\eqref{eq:interp} is approximately satisfied for examples from a training data set. The same should also hold for examples from a validation data set, which is not used to optimize the parameters.
The training objective is commonly modeled as an expected loss to be minimized: 
\begin{equation}
\label{eq:opt}
    \min_{\bftheta}  \Phi(\bftheta), \quad \text{ where } \quad \Phi(\bftheta) =  \frac1s \sum_{k=1}^s {\rm loss}(f(\bfy^k,\bftheta), \bfc^k) + R (\bftheta).
\end{equation}
Here, $(\bfy^1,\bfc^1), \ldots, (\bfy^s,\bfc^s) \in \mathcal{Y}\times\mathcal{C}$ are the training data, $f(\bfy,\bftheta)$ includes the action of the neural network, but may also contain other layers, e.g., fully-connected layers, opening and connective layers, and softmax transformations in classification. $R$ is a regularization term.
The optimization problem in Eq.~\eqref{eq:opt} is typically solved with gradient-based non-linear optimization techniques~\cite{bottou2018optimization}. Due to the large scale and nature of the learning problem, it is common to use variants of stochastic gradient descent (SGD) like Adam~\cite{kingma2014adam} to minimize \eqref{eq:opt}.
These methods perform iterations using gradient information from randomly chosen subsets of the data.

\subsection{\textbf{Quantization-Aware Training}}
\label{sub:background_quant}
As mentioned before, in this paper we focus on quantized neural networks, and in particular, on the scenario of quantization-aware training. We restrict the values of the weights to a smaller set, so that after training, the calculation of a prediction by the network can be carried out in fixed-point integer arithmetic. The intermediate hidden layers are quantized (rounded) as well. Even though the quantization involves a discontinuous rounding function, the discrete weights can be optimized using gradient-based methods, also known as quantization-aware training schemes~\cite{han2016deep,yin2019blended}. During the forward pass, both the weights and hidden activation maps are quantized, while during gradient calculation in the backward pass, derivative information is passed through the rounding function, whose exact derivative is zero. This method is known as the Straight Through Estimator (STE)~\cite{yoshuabengio2013}. We note that the gradient-based optimization for the weights is applied in floating-point arithmetic as we describe next, but during inference, the weights and activations are quantized, and all operations are performed using integers only.

\begin{figure}
\begin{center}
\begin{subfigure}[b]{0.32\textwidth}
         \centering
         \includegraphics[width=\textwidth]{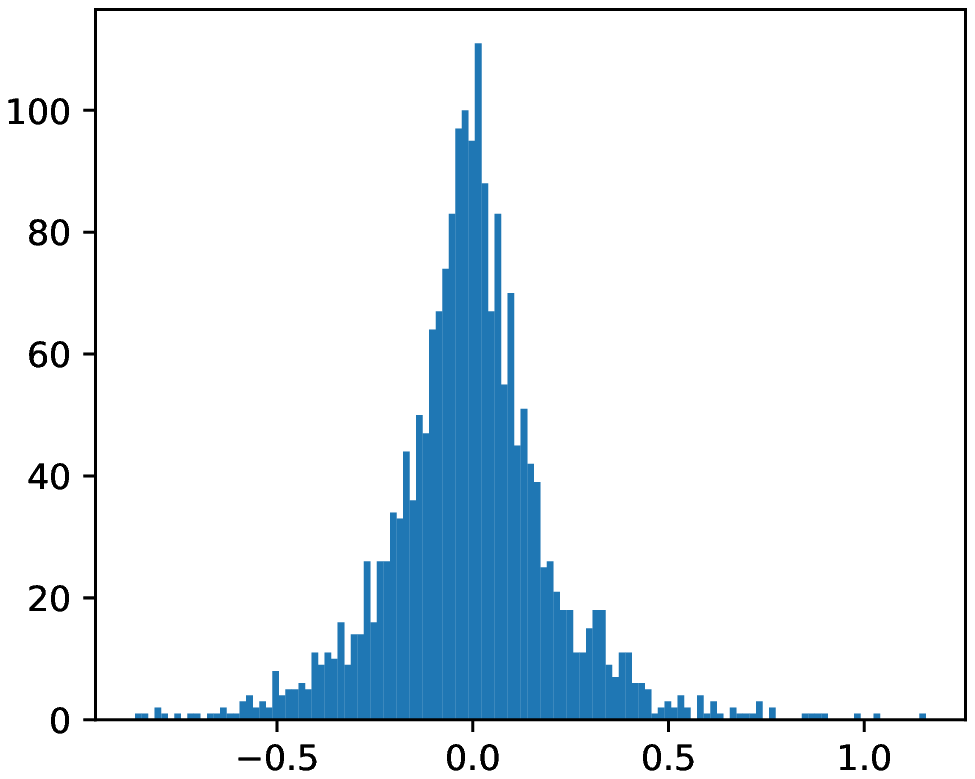}
         \caption{Original signal}
     \end{subfigure}
     \hfill
     \begin{subfigure}[b]{0.32\textwidth}
         \centering
         \includegraphics[width=\textwidth]{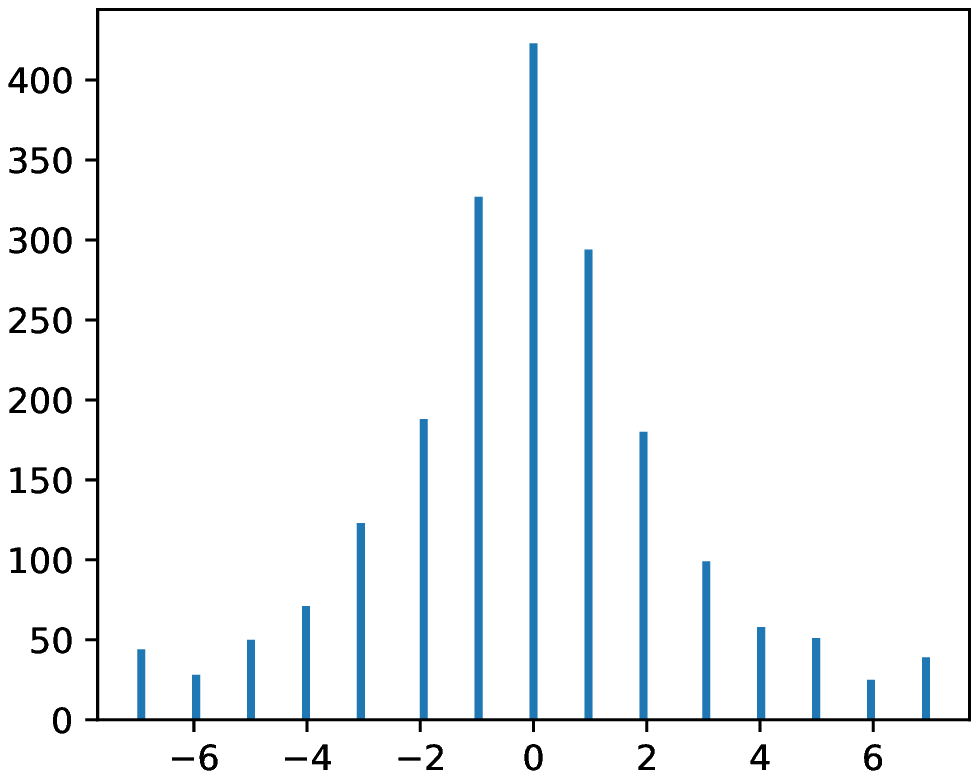}
         \caption{Quantized to integers}
     \end{subfigure}
     \hfill
     \begin{subfigure}[b]{0.32\textwidth}
         \centering
         \includegraphics[width=\textwidth]{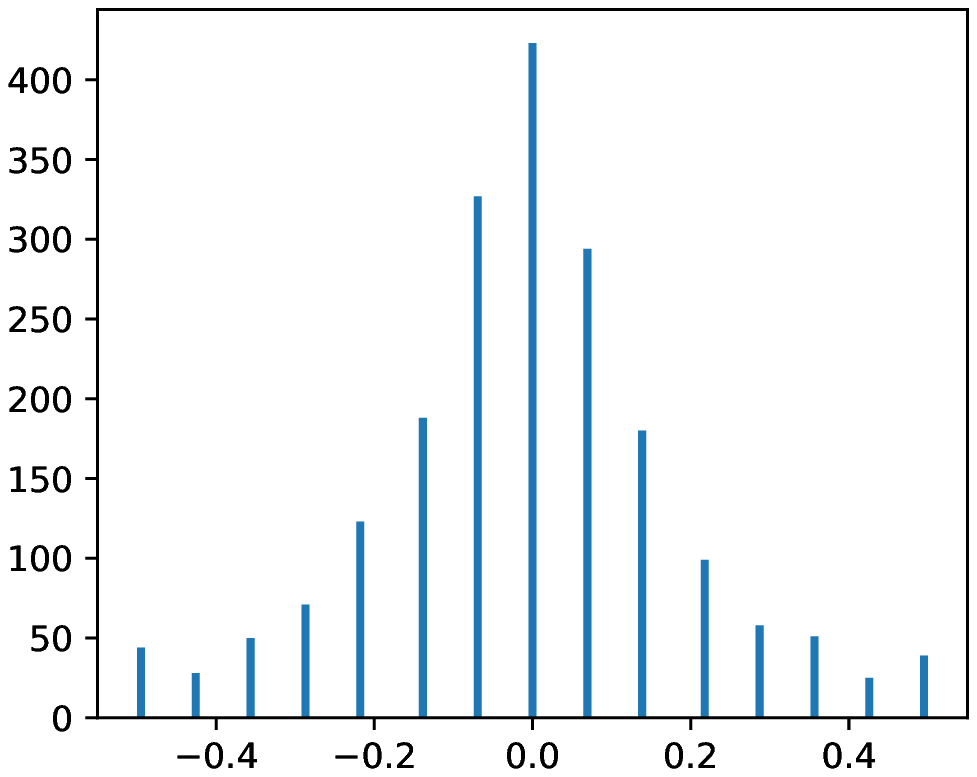}
         \caption{Floating-point values}
     \end{subfigure}
\caption{An example of a signal that has been uniformly quantized to 4 bits per sample. The original histogram is given in (a). The values are then clipped to the range $[-\alpha,\alpha]$ (in this example $\alpha=0.5$), multiplied by $7/\alpha$ ($7=2^{4-1}-1$, for 4 bits), quantized to integer values in (b), and scaled back to their original values in (c) by multiplying with the reciprocal $\alpha/7$.}
\label{fig:quantized_weights_example}
\end{center}
\end{figure}

We now present the details of the quantization scheme that we use, based on~\cite{li2019apot}. First, we define the pointwise quantization operator:
\begin{equation}
\label{eq:quantoperator}
    q_b(t) = \frac{\mbox{round}((2^b - 1) \cdot t)}{2^b - 1},
\end{equation}
where $t$ is a real-valued scalar in one of the ranges [-1, 1] or [0, 1] for signed or unsigned quantization\footnote{We assume that the ReLU activation function is used in between any convolution operator, resulting in non-negative activation maps, and can be quantized using an unsigned scheme. If a different activation function is used that is not non-negative, like $\tanh()$, signed quantization should be used instead.}, respectively. $b$ is the number of bits that are used to represent $t$ as integer at inference time (during training, $t$ and $q_b(t)$ are real-valued). During the forward pass, we force each quantized value to one of the ranges above by applying a clipping function to the quantized weights and activations before applying Eq. \eqref{eq:quantoperator}: 
 \begin{eqnarray} \label{eq:quantweights}
 w_b &=& Q_b(w) =  \alpha_w q_{b-1}\left(\mbox{clip}\left(\frac{w}{\alpha_w}, -1, 1\right)\right) \nonumber \\
x_b &=& Q_b(x) =  \alpha_x q_{b}\left(\mbox{clip}\left(\frac{x}{\alpha_x}, 0, 1\right)\right).
\end{eqnarray}
 Here, $w, w_b$ are the real-valued and quantized weight tensors, $x, x_b$ are the real-valued and quantized input tensors, and $\alpha_w, \alpha_x$ are their associated clipping parameters, respectively. An example of Eq. \eqref{eq:quantweights} applied to a weight tensor using 4-bit signed quantization, is given in Fig. \ref{fig:quantized_weights_example}. During training, we iterate on the floating-point values of the weights $w$, while both the weights and activation maps are quantized in the forward pass (i.e., $x_b$ and $w_b$ are passed through the network). The STE~\cite{yoshuabengio2013} is used to compute the gradient in the backward pass, where the derivative of $q_b$ is ignored and we use the derivatives w.r.t $w_b$ in the SGD optimization to update $w$ iteratively. 

The quantization scheme in Eq. \eqref{eq:quantweights} involves the clipping parameters $\alpha_w$ and $\alpha_x$, also called scales, that are used to translate the true value of each integer in the network to its floating-point value. This way, at inference time, two integers from different layers can be multiplied in a meaningful way that is faithful to the original floating-point values~\cite{gemmlowp,hubara2017quantized}. These scales also control the quantization error, and need to be chosen according to the values propagating through the network and the bit allocation $b$, whether the weights or the activations. As shown in~\cite{soudry1}, the quantization error can be evaluated by
\begin{equation}\label{eq:MSE}
\mbox{MSE}(\bfx,\bfx_b) = \frac{1}{n}\sum_{i=1}^{n} (x_{i}-(x_b)_{i})^2,
\end{equation}
and it is reasonable to reduce the MSE to maximize the accuracy of the quantized network. In particular, the scales $\alpha_w,\alpha_x$ can be computed by minimizing the MSE based on sampled batches of the training data~\cite{soudry1}.

Alternatively, the works~\cite{li2019apot,esser2019learned}, which we follow here, introduced an effective gradient-based optimization to find the clipping values $\alpha_x,\alpha_w$ for each layer. Given Eq. \eqref{eq:quantweights} the gradients w.r.t $\alpha_w$, and $\alpha_x$ can be approximated using the STE~\cite{li2019apot,esser2019learned}. For the activation maps, for example, this resolves to:
\begin{eqnarray}
\label{eq:act_alpha_derivative}
    \frac{\partial{x_b}}{\partial{\alpha_x}} = 
    \begin{cases}
    0 & \text{if $x \leq 0$} \\
    1 & \text{if $x \geq \alpha_x$} \\
    \frac{x_b}{\alpha_x} - \frac{x}{\alpha_x} & \text{if $0 < x < \alpha_x$}. \\
    \end{cases}
\end{eqnarray}
This enables the quantized network to be trained end-to-end with backpropagation.
To further improve the optimization,~\cite{li2019apot} normalize the weights before quantization:
 \begin{equation} \label{eq:normalize}
     \hat w = \frac{w - \mu}{\sigma  + \epsilon}.
\end{equation}
Here, $\mu$ and $\sigma$ are the mean and standard deviation of the weight tensor, respectively, and $\epsilon=10^{-6}$.

\subsection{\textbf{Motivation: the Importance of Stability in Quantized CNNs}}
\label{sub:background_motivation}
Neural networks are known to be susceptible to noise in their inputs, in the sense that a small perturbation to the input image can cause a trained classifier to assign an incorrect label. This phenomenon can easily occur for examples $\bfy$ that are sufficiently close to the decision boundary. For these examples, a small perturbation is likely to change the resulting label. Similarly, the quantization of the activation maps inevitably adds noise to the input of \emph{every} layer. While the quantization error is not chosen specifically for a given input, it does appear in every layer. Hence, it can potentially be as harmful as an adversarial attack with regards to the output and accuracy of the network. The introduction of such errors throughout the layers of the network resembles the approximation and round-off errors in time integration of PDEs which require forward stability to converge. To relieve this phenomenon, we follow the approach of constructing stable architectures that are more robust against such errors~\cite{HaberRuthotto2017}.

We call a discrete forward propagation like Eq.~\eqref{eq:resnetintro} \textit{stable} if it prevents any perturbation from growing as it propagates through time steps. More explicitly, the $N$-layered network in Eq. \eqref{eq:resnetintro} is stable if there exists some $M>0$ such that
$$
\|\bfx_N-\tilde\bfx_N\| \leq M \|\bfx_i-\tilde\bfx_i\|, \quad i=1,...,N-1
$$
where $\bfx_i$ and $\tilde\bfx_i$ are the true and perturbed feature maps of the $i$-th layer, respectively. The Jacobian of $\bfx_N$ with respect to $\bfx_i$ can then be bounded proportionally to $M$, and accordingly, any stable architecture needs to have a bounded Jacobian~\cite{ruthotto2019deep}. Similar arguments regarding regularization for promoting bounded Jacobians were applied in~\cite{jakubovitz2018improving} to withstand adversarial attacks. In this work we focus on the symmetric ResNet in Eq. \eqref{eq:resnet} and \eqref{eq:doubleSym}. Indeed, under mild assumptions on $\sigma$, the Jacobian of the symmetric ResNet layer is bounded by 1 if the norm of $\bfK$ is small enough, and the network is stable~\cite{ruthotto2019deep}.  

\section{Treating Activation Error as Noise Using Total Variation}
\label{sec:tv_cnn}
The quantization scheme in Eq. \eqref{eq:quantweights} is applied to both the weights and activation (feature) maps throughout the network. The weights are known during training and are fixed during inference. Thus, their post-quantization behaviour is known and we can regularize them during training. On the other hand, the activations are available only at inference time and depend on the input data. Therefore, they are prone to erratic behaviour given unseen, possibly noisy data. For this reason, we view the difference $\eta_i = \bfx_i - (\bfx_b)_i$ at at the i-th layer  as noise generated from the quantization of the activations. In what follows, we propose a method to design a network that prevents the amplification of this noise.
    
\subsection{\textbf{Total Variation in Quantized Networks}}
One of the most popular and effective approaches for image denoising is the Total Variation method suggested by Rudin, Osher, and Fatemi~\cite{RudinOsherFatemi1992}. This approach seeks to regularize the data loss with the TV norm of the image, 
\begin{equation}\label{eq:TV}
||u||_{TV(\Omega)}  = \int_{\Omega}{\|\nabla u(x)\| dx} 
\end{equation}
where as a norm $\|\cdot\|$ one can use $\ell_2$ or $\ell_1$, which is the applied to the gradient of the image. In this work we consider the $\ell_1$ norm over $\ell_2$ because of its lower computational cost (more details later).

\begin{figure}
\begin{center}
\begin{subfigure}[b]{0.48\textwidth}
         \centering
         \includegraphics[width=\textwidth]{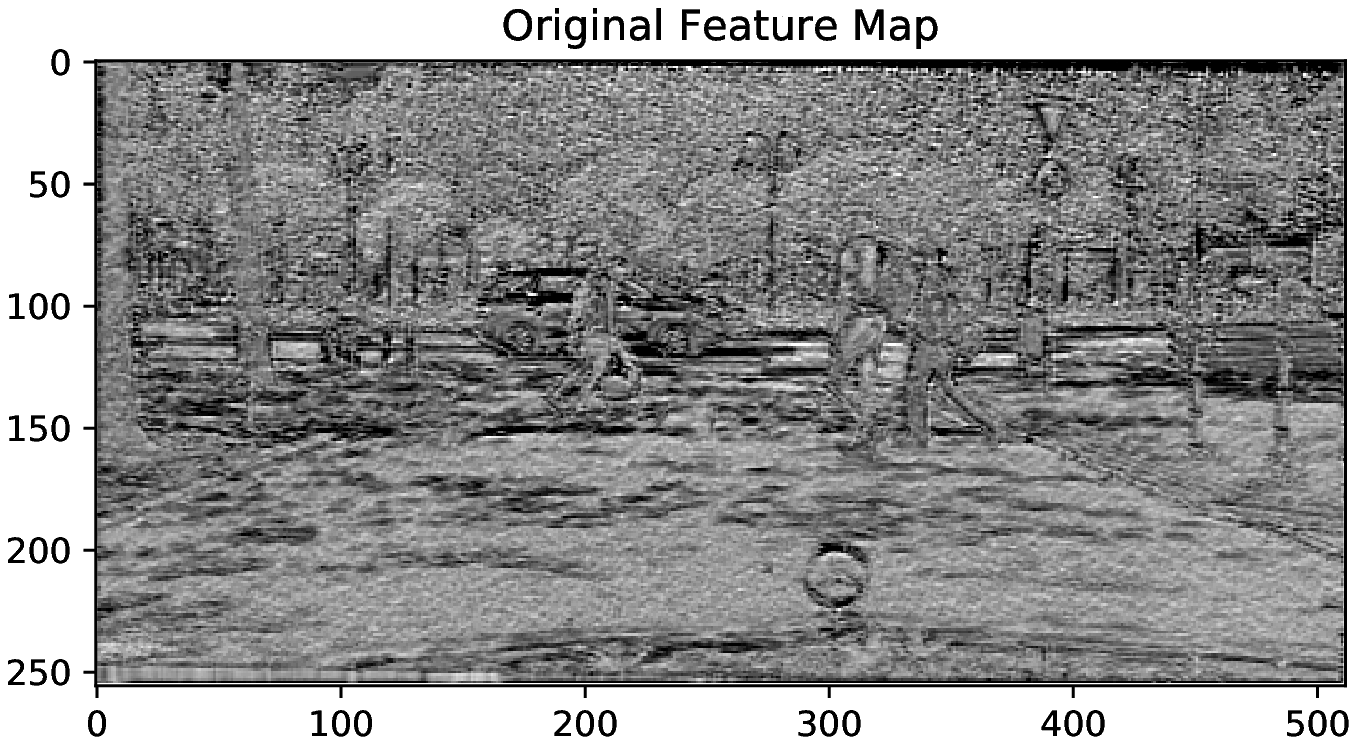}
         \caption{Original feature map}
     \end{subfigure}
     \hfill
     \begin{subfigure}[b]{0.48\textwidth}
         \centering
         \includegraphics[width=\textwidth]{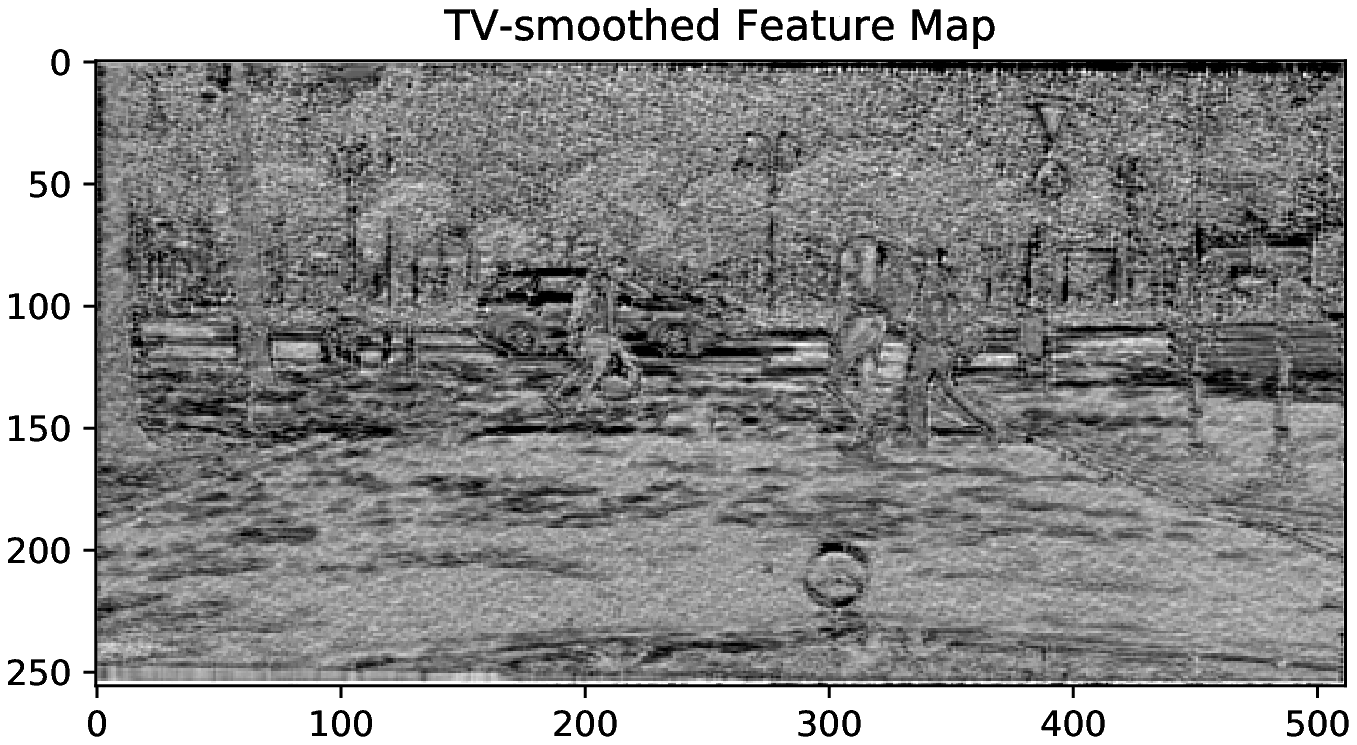}
         \caption{TV-smoothed feature map}
     \end{subfigure}
     
     \bigskip
     
     \begin{subfigure}{\linewidth}
        \centering
        \begin{tikzpicture}
          \begin{axis}[
                ymode=log,
                width=\linewidth,
                height=0.5\linewidth,
                legend style={draw=none,cells={anchor=west}},
                legend columns=1,
                xtick={-4,-3,-2,-1,0,1,2,3,4},
                every axis plot post/.style={thick},
                domain=-4:3.5
            ]
            
            \addplot[blue,name path=orig,domain=-4:3.5]
            table[x=x,y=y,col sep=comma] {data/tvsmooth_orig.csv};
            
            \addplot[red,name path=smooth,domain=-4:3.5]
            table[x=x,y=y,col sep=comma] {data/tvsmooth.csv};
            
            \legend{Original, TV-smoothed}
          \end{axis}
        \end{tikzpicture}
        \caption{Original distribution  $\in [-3.78,3.11]$ vs TV-smoothed distribution $\in [-2.05, 1.24]$}
    \end{subfigure}
     \hfill
\caption{An example of a feature map from the 4-th layer of the ResNet50 encoder for an image. (a) and (b) show the feature map before and after 3 iterations of the TV-smoothing operator in Eq. \eqref{eq:smooting_layer}, with $\gamma^2=0.1$. It is evident that fine details are preserved after the rather light smoothing. (c) shows the corresponding value distributions. It is evident that the TV-smoothing eliminates outliers in addition to denoising the image a bit. After a 4-bit signed quantization with $\alpha = 0.5$, the MSE between the original and quantized maps are 0.16 and 0.05 for the original and TV-smoothed maps, respectively. Hence, we see that the TV-smoothing outputs a feature map that is better suited for quantization. The distributions in (c) have been smoothed slightly for improved visual clarity.}
\label{fig:TV}
\end{center}
\end{figure}

We wish that the feature maps to have a low TV norm, and to this end we minimize it using regularization during training. Since a low TV norm is known to encourage piecewise constant images, we can incur less errors when quantizing these images. This is because they will typically contain less gray-scale values. Furthermore, extreme values influence the MSE in Eq. \eqref{eq:MSE} the most, and one will typically choose a smaller clipping parameter $\alpha_x$ as the number of bits decreases, to better approximate the distribution of values by the quantized distribution~\cite{soudry1}. Since both of these factors negate each other, we wish to eliminate extreme values (outliers) from the distribution as much as possible, such that that the optimization of $\alpha_x$ can better fit the distributions. Smoothing the feature maps essentially averages pixels, which promotes more uniform distributions and reduces outliers, as demonstrated in Fig. \ref{fig:TV}. 

However, in our experience, regularizing solely towards smooth feature maps also harms the performance of the network, since it de-emphasizes other important parts of the loss function during training. We have witnessed experimentally that following an aggressive application of the TV regularization in addition to the loss function, the $3\times3$ filters in the network gravitate towards learning low-pass filters rather than serving as spatial feature extractors, like edge detectors. To remedy this, we add TV smoothing layers to network, acting as non-pointwise activation layers. This addition results in piece-wise smooth feature maps with significantly less noise in the feature maps, allowing other filters in the network to focus on feature extraction and minimization of the loss function. As we show next, this can be done at a  lower cost compared to the convolution operations.
  
In practice, in every non-linear activation function, we also apply an edge-aware denoising step. The denoising step is applied as
\begin{equation}\label{eq:smooting_layer}
S(\bfx) = \bfx - \gamma^2(\bfD_x + \bfD_y)\bfx,
\end{equation}
and instead of using a $\sigma(x)$ activation, we apply $\sigma(S(x))$ as a non-pointwise activation. The operators $\bfD_x$ and $\bfD_y$ in \eqref{eq:smooting_layer} are the weighted Laplacians in the $x$ and $y$ directions, respectively, that smoothly approximate the TV norm. For example $\bfD_x = \bfG_x^T\bfW\bfG_x$ where $\bfW = \mbox{diag}\{(|\bfG_x\bfx|+\epsilon )\}^{-1}$.
The operators $\bfG_x$ and $\bfG_y$ are standard binary 1D convolution operators with the kernels $[-1,1]$ and $[-1,1]^T$ that do not involve any multiplications (only additions and subtractions), and can be therefore applied efficiently in-place given a window of values. At inference time, only the sign of $\bfG_x\bfx$ is required for the computation of $\bfW$, resulting in computationally efficient operators. These operators are hard-coded and fixed, and not learnt, meaning they do not incur memory overhead, as opposed to standard learnt depthwise convolutions. We note that in the choice of $\bfD_x$ and $\bfD_y$ above, often referred to as ``anisotropic TV'', the filters are separated in each direction. An alternative that may perform slightly better for denoising can be the use of $\bfW = \mbox{diag}\{(\sqrt{(\bfG_x\bfx)^2 + (\bfG_y\bfx)^2 + \epsilon^2})\}^{-1}$, while considering the additional computational cost of this more evolved operations. Here we consider the anisotropic version so that the computational overhead is as low as possible. The parameter $\gamma$ is the only learnt parameter in Eq. \eqref{eq:smooting_layer}, which is a single parameter per layer, and is essential for the network to adapt to the amount of smoothing that needs to be applied at each step. This step also adds some non-linearity to the network, and for a small enough $\gamma$, promotes stability. Lastly, we note that the example in Fig. \ref{fig:TV} is applied using three TV layers, while in our network we apply one TV layer in every step. Indeed, a single TV step has a lesser effect than three, but throughout the forward application of our network, the smoothness of the feature maps accumulates with the layers, leading to a similar effect of the advanced feature maps.  

\section{Stable and Quantized Residual Networks}
\label{sec:stable_nets}
As discussed earlier, our goal is to promote stability in CNNs, to prevent the amplification of quantization errors. To obtain that, we would like the quantized network to behave similarly (in terms of its activations) to a non-quantized instance of the same network. The symmetric variant of ResNet in Eq. \eqref{eq:resnet} and \eqref{eq:doubleSym}, together with the activation quantization operator in Eq. \eqref{eq:quantweights} is given as follows:
\begin{equation}
\label{eq:stable_quant_sym_resnet}
    \bfx_{j+1} = Q_b(\bfx_j - h \bfK_j^\top\ Q_b(\sigma(\bfK_j\bfx_j))),
\end{equation}
where $Q_b$ denotes the quantization operation in Eq. \eqref{eq:quantweights}.
Let us denote the feature maps of the non-quantized network by $\hat\bfx_j$ (i.e., assuming $q_b(x) = x$ in Eq. \eqref{eq:quantweights}), and let the error between corresponding activations be denoted by $\eta_j = \bfx_j - \hat\bfx_j$. We consider the same weights $\bfK$ between the two architectures, whether quantized or not, and also assume that the absolute quantization error for any scalar is bounded by some $\delta$. We analyze the propagation of the quantization error $\eta_{j+1}$ as a function of $\eta_{j}$.

Let us first assume that $\sigma$ is Lipschitz continuous such that 
\begin{equation}
\label{eq:Lipschitz}
    \vert \sigma(x_1) - \sigma(x_2) \vert \leq L \vert x_1 - x_2 \vert
\end{equation}
holds for some real $L \geq 0$ and for all real $x_1, x_2$. Also assume that $\sigma$ is monotonically non-decreasing, hence \eqref{eq:Lipschitz} also means that for every $x_1,x_2$ we have 
\begin{equation}
\label{eq:monotone_Lipschitz}
    \sigma(x_1) - \sigma(x_2)  = \omega (x_1 - x_2),\quad  \omega\in [0, L],
\end{equation}
which is obtained because the signs in \eqref{eq:Lipschitz} are consistent on both sides following the monotonicity. We note that the ReLU activation that we use here satisfies \eqref{eq:monotone_Lipschitz} for $L=1$.

We start unwrapping the block in Eq. \eqref{eq:stable_quant_sym_resnet}. First, we subtract $\hat\bfx_{j+1}$ from both sides, and replace the outer quantization with the error term $\eta_{j_1}$. 
\begin{equation}
    \eta_{j+1} = \eta_{j_1} + \bfx_j - h \bfK_j^\top Q_b(\sigma(\bfK_j\bfx_j)) - \hat\bfx_{j+1}
\end{equation}
Next, we remove the other instance of $Q_b$ and add another error term $\eta_{j_2}$:
\begin{equation}
    \eta_{j+1} = \eta_{j_1} -h\bfK_j^\top\eta_{j_2} +  \bfx_j - h \bfK_j^\top\sigma(\bfK_j\bfx_j) - \hat\bfx_{j+1}.
\end{equation}
Remembering that $\bfx_j = \hat\bfx_j+\eta_j$, then using \eqref{eq:monotone_Lipschitz} we have
\begin{equation}
    \sigma(\bfK_j\bfx_j) = \sigma(\bfK_j\hat\bfx_j) + \bfOmega\bfK_j\eta_j,
\end{equation}
where $\bfOmega$ is a diagonal matrix for which $\bfOmega_{ii}\in[0,L]$.
Therefore, we have

\begin{eqnarray}
\label{eq:error_prop}
    \eta_{j+1} &=& \eta_{j_1} -h\bfK_j^\top\eta_{j_2} +  \hat\bfx_j+\eta_j - h \bfK_j^\top(\sigma(\bfK_j\hat\bfx_j) + \bfOmega\bfK_j\eta_j) - \hat\bfx_{j+1} \nonumber\\
    & = & \eta_{j_1} -h\bfK_j^\top\eta_{j_2} + \eta_j -h\bfK_j^\top\bfOmega\bfK_j\eta_j  \nonumber \\
    & = & (\bfI -h\bfK_j^\top\bfOmega\bfK_j)\eta_j + \eta_{j_1} -h\bfK_j^\top\eta_{j_2}.
\end{eqnarray}

The key ingredient of the analysis above is that $\eta_{j_1}$ and $\eta_{j_2}$ are fixed and bounded for every layer. On the other hand, it is the matrix 
$\bfJ_j = \bfI -h\bfK_j^\top\bfOmega\bfK_j$
that multiplies $\eta_j$ at every iteration and propagates the previous error into the next block. 
Since $\bfOmega_{ii} \geq 0$, it means that 
$\bfK_j^\top\bfOmega\bfK_j$ is positive semi-definite, and with a proper choice of $\bfK_j$ and $h$, we can force $\rho(\bfJ_j)<1$, so that the error decays. To ensure this forward stability we must set $h < 2(L\|K_j\|_2^2)^{-1}$ for every layer $j$ in the network, where $L$ is the upper bound for $\sigma'()$~\cite{alt2021translating} (the Lipschitz constant of $\sigma$). This is generally possible in ResNets only if we use the symmetric variant in Eq. \eqref{eq:doubleSym}. ~\cite{ruthotto2019deep,zhang2020forward} also achieved a similar stability result, but not in the context of quantization. 

\subsection{\textbf{Stable Channel and Resolution-Changing Layers}}
\label{sub:stable_connector}
The classical CNN architecture typically applies several layers like Eq. \eqref{eq:resnetintro}. However, to obtain better representational capabilities for the network, the number of channels often increases every few layers. In CNNs, this is sometimes accompanied by a down-sampling operation performed by a pooling layer or a strided convolution. In such cases, the residual equation Eq. \eqref{eq:resnetintro} cannot be used, since the update $F(\bfx_j,\bftheta_j)$ does not match $\bfx_j$ in terms of dimensions. For this reason, ResNet authors sometimes favour including 3-4 steps like
\begin{equation}
    \bfx_{j+1} = \bfK_j\bfx_j + F(\bfx_j,\bftheta_j),\quad \mbox{or}\quad    \bfx_{j+1} = F(\bfx_j,\bftheta_j), 
\end{equation}
throughout the network to match the changing resolution and number of channels. These layers are harder to control than Eq. \eqref{eq:resnetintro} and \eqref{eq:doubleSym} if one wishes to ensure stability.   

In this work we are interested in testing and demonstrating the theoretical property described in section \ref{sec:stable_nets} empirically, using fully stable networks (except for the very first and last layers). To this end, when increasing the channel space from $n_{c_{in}}$ to $n_{c_{out}}$ for $\bfx_{j+1}$ and $\bfx_j$ respectively, we simply concatenate the output of the step with the necessary channels from the current iteration to maintain the same dimensions. That is, we apply the following:
\begin{equation}
\label{eq:stable_connector}
    \bfx_{j+1} = \begin{bmatrix}
    \bfx_j + h F(\bfx_j, \bftheta_{j})\\
    (\bfx_j)_{1:n_{c_{out}} - n_{c_{in}}}
    \end{bmatrix},
\end{equation}
where $(\bfx_j)_{1:n_{c_{out}} - n_{c_{in}}}$ are the first $n_{c_{out}} - n_{c_{in}}$ channels of $\bfx_j$. This assumes that each of the channel changing steps satisfies $n_{c_{in}} \leq n_{c_{out}} \leq 2n_{c_{in}}$, which is quite common in CNNs. This way, we only use the symmetric dynamics as in Eq. \eqref{eq:doubleSym} throughout the network, which is guaranteed to be stable for a proper choice of parameters. Finally, we do not apply strides as part of Eq. \eqref{eq:stable_connector}, and to reduce the channel resolution we simply apply average pooling following Eq. \eqref{eq:stable_connector}, which is a stable, parameter-less, operation. 

\subsection{\textbf{A Stable Variant of MobileNetV2}}
\label{sub:mobilenet}
In addition to standard residual networks, we also consider the popular MobileNet family of light-weight CNN architectures~\cite{sandler2018mobilenetv2,howard2019searching}, which are the most common architectures for edge devices, achieving very nice results while requiring modest computational resources to deploy. These architectures utilize the ``inverse bottleneck'' structure, where the channel space throughout the network is relatively small, but in every step it is expanded and reduced by $1\times1$ convolutions, with ``depthwise'' $3\times 3$ convolutions in the middle expanded channel space. The depthwise convolutions apply $3\times3$ kernels on each channel without mixing between the channels, and hence they are less expensive than $1\times1$ convolutions. The general structure of MobileNetV2~\cite{sandler2018mobilenetv2} reads:
\begin{equation}
\label{eq:mobilenet}
    \bfx_{j+1} = \bfx_j + \bfK_{3, j}\sigma(\bfK_{2, j}\sigma(\bfK_{1, j}\bfx_j))
\end{equation}
where $\bfK_{3, j},\bfK_{1, j}$ are two different learnable $1 \times 1$ convolution filters, and $\bfK_{2, j}$ is a learnable depthwise $3 \times 3$ convolution matrix. Due to the relatively small channel space used throughout the architecture, the quantization of this network to low bit-rates is challenging.   

The MobileNetV2 architecture in Eq. \eqref{eq:mobilenet} can also be seen as a residual network, and is generally unstable. Because of its inverse bottleneck structure, it does not fit the structure of Eq. \eqref{eq:resnetintro} and Eq. \eqref{eq:doubleSym}. To test the importance of stability under quantization for such networks as well, we define the stable MobileNetV2 variant:
\begin{equation}
\label{eq:stableMobileNet}
    \bfx_{j+1} = \bfx_j - (\bfK_{1, j})^\top((\bfK_{2, j})^\top\sigma (\bfK_{2, j}\bfK_{1, j}\bfx_j)),    
\end{equation}
where now, the separable convolution operator $\bfK_{2, j}\bfK_{1, j}$ takes the role of the single operator $\bfK$ in Eq. \eqref{eq:doubleSym}. Going between Eq. \eqref{eq:mobilenet} and \eqref{eq:stableMobileNet} we lost one non-linearity, but as we show later, this has marginal influence on the accuracy of the network. We also apply the depthwise operation twice with the same kernel weights, but this is inexpensive compared to $1\times1$ convolutions. Changing the resolutions and channel space is done the same way as described in section \ref{sub:stable_connector}.  

\subsection{\textbf{Stable and Quantized GCNs}}
\label{sub:stable_gcns}
To show that stability is useful beyond CNNs operating on structured data, we also study the importance of stability for quantized Graph Convolution Networks (GCNs), which can be thought of as a PDE discretized on unstructured grids~\cite{eliasof2020diffgcn}. Specifically, we use the diffusive PDE-GCN architecture formulated in~\cite{eliasof2021pdegcn}, which utilizes symmetric operators as in Eq. \eqref{eq:doubleSym}, only on unstructured graphs:
\begin{equation}
\label{eq:stableDiffusiveGCN}
    \bfx_{j+1} = \bfx_j - h \bfS_j^{\top} \bfK_j^{\top} \sigma(\bfK_j \bfS_j \bfx_j).
\end{equation}
Here, $\bfx _j$ are the features defined over the nodes of a graph, $\bfK_j$ is a learnt $1\times1$ convolution operator and $\bfS_j$ is either learnt (e.g., as in~\cite{eliasof2020diffgcn}) or pre-defined spatial operation (e.g., the graph Laplacian), both for the $j$-th layer of the network. Here, the forward stability is guaranteed in the continuous case by the symmetry of the operator~\cite{eliasof2021pdegcn}. 

We refer to a network that is driven by the dynamics in Eq. \eqref{eq:stableDiffusiveGCN} as PDE-GCN\textsubscript{D}(sym.). Analogously, we define the non-symmetric residual layer:
\begin{equation}
\label{eq:unstableDiffusiveGCN}
    \bfx_{j+1} = \bfx_j - h \bfS_j^{\top} \bfK_{j_2} \sigma(\bfK_{j_{1}} \bfS_j \bfx_j)
\end{equation}
Where $\bfK_{j_{1}}$ and $\bfK_{j_{2}}$ are distinct learnt $1\times 1$ convolution operators. This formulation does not necessarily yield a symmetric operator. Therefore we denote a network that is governed by such dynamics by PDE-GCN\textsubscript{D}(non-sym.). The quantization for both Eq. \eqref{eq:stableDiffusiveGCN} and \eqref{eq:unstableDiffusiveGCN} is applied to the weights and before each of the combined convolution operators. 

\section{Numerical Experiments}
\label{sec:experiments}
To validate our propositions, we apply our formulations to several learning tasks and data-sets, from image classification and segmentation (using CNNs) to graph node classification (using GCNs).

\subsection{\textbf{Settings and Data-sets}}
\label{sub:datasets}
Our code is implemented in PyTorch, and all experiments are conducted on an Nvidia RTX-3090 with 24GB of memory. Below, we elaborate on the different data-sets explored throughout the numerical experiments.

\noindent\textbf{CIFAR-10/100.}
The CIFAR-10/100 image classification data-sets~\cite{krizhevsky2009learning} each consist of 60k natural images of size $32\times32$ where each image is assigned to one of ten categories (for CIFAR-10) or one hundred categories (for CIFAR-100). The data-set includes 50K training examples and 10K test examples. We derive a validation set for training by holding out 10\% of the training data and report accuracy metrics on the test data.
\\
\noindent\textbf{ImageNet.}
The ImageNet~\cite{ImageNet} ILSVRC 2012 challenge consists of $1.28$M training images and $50$K validation images from a thousand categories. As in~\cite{he2016deep},  we resize the images to $224\times 224$ and use standard data augmentation of random horizontal flips~\cite{he2016deep}.
\\
\noindent\textbf{Cityscapes.}  The image semantic segmentation data set~\cite{cordts2016cityscapes} contains 5000 finely-annotated images with 19 categories ranging from road and vehicles to trees and people. We use the standard train-validation data split as in~\cite{cordts2016cityscapes} , i.e.; 2975 and 500 for training and validation, respectively.
In addition, we use standard augmentations like random horizontal flips.
\\
\noindent\textbf{Node-classification data-sets.}
Lastly, we use graph neural networks on three citation network data-sets: Cora, CiteSeer and PubMed~\cite{sen2008collective}. The statistics of the data-sets can be found in Tab. \ref{tab:semisupervised_statistics}. For each data-set, we use the standard train/validation/test split as in~\cite{yang2016revisiting}, with 20 nodes per class for training, 500 validation nodes and 1,000 testing nodes. We use the same training scheme as in~\cite{chen2020simple}. 

\subsection{\textbf{Total Variation Activation Functions}}
\label{sub:experiment_tv}
In this section we add the TV activation function to popular architectures for common tasks, without any additional changes. This is the most common use-case for quantization methods, where we apply quantization to existing architectures, usually given pre-trained weights. 

\begin{table}
    \centering
\begin{tabular}{c c c c}
    \toprule
\bfseries Model & \bfseries Method & \bfseries 3W/3A & \bfseries 4W/4A \\ 
\midrule
        ResNet50    & DoReFa-Net~\cite{drfn}      & 69.9                  & 71.4 \\ 
        FP 76.4\%   & LQ-Nets~\cite{zhang2018lqnets}        & 74.2                  & 75.1 \\ 
                  & PACT~\cite{choi2018pact}            & 75.3                  & 76.5  \\ 
                  & SAT~\cite{jin2020scaleadjusted} & 75.3                 & 76.3\\
                  & TV (ours)    & 75.4                 & 76.2 \\
    \bottomrule
\end{tabular}
    \caption{Image classification results using our TV layers on ImageNet.}
    \label{tab:ImageNetTV}
\end{table}

\begin{table}
    \centering
\begin{tabular}{c c c c c}
    \toprule
\bfseries Model & \bfseries Method & \bfseries 2/2  & \bfseries 3/3 & \bfseries 4/4 \\ 
\midrule
        ResNet20      &   DoReFa-Net~\cite{drfn}                 &   -       &   68.4    &   68.9 \\
        FP 70.35\%    &   LQ-Nets~\cite{zhang2018lqnets} &   -       &   68.4   &    69.0 \\
                    &   WNQ~\cite{wnq}                 &   -       &   68.9    &   69.0 \\
                    &   TV (ours)                     &   68.12   &   70.32   &   70.34 \\
    \bottomrule
\end{tabular}
    \caption{Image classification results using our TV layers on CIFAR-100.}
    \label{tab:Cifar100TV}
\end{table}

First, we consider the image classification task on the CIFAR-100 and ImageNet data-sets, where we add the TV layers to quantized CNNs. Our post-training procedure is comprised of two phases. The first phase begins with a pre-trained network and trains it to the desired number of bits as described in section \ref{sub:background_quant}, based on the work of~\cite{zhou2018adaptive}. In the second phase, the network is augmented with additional TV layers. The $\gamma$ parameters described in Eq. \eqref{eq:TV} are learnt as part of the training, and control the strength of the edge-aware smoothing that we add to the network. We observe that the learnt $\gamma$ parameters are kept low, such that our TV layers promote stability to the network, as the norm of their Jacobian is smaller than 1. The results on ImageNet are reported in Tab. \ref{tab:ImageNetTV}, where the baseline model with float-precision reads 76.4 $\%$. We show, that our method is superior or on par with other popular methods such as DoReFa~\cite{drfn} LQ-Nets~\cite{zhang2018lqnets}, PACT~\cite{choi2018pact} and SAT~\cite{jin2020scaleadjusted}. In addition, Tab. \ref{tab:Cifar100TV} provides the classification accuracy on CIFAR-100. There, the gain of our TV layers is more significant. For instance, while recent methods like LQ-Nets and WNQ achieve an accuracy of 68.9 $\%$ and 68.4$\%$ with 3 weight and activation bits, respectively, ours obtains 70.32$\%$. 

\begin{table}
    \centering
\begin{tabular}{lccc}
    \toprule
\bfseries Method & \bfseries Precision & \bfseries Acc. ($\%$) &  \bfseries mIoU \\
\midrule
    ResNet-DeepLab  & 32W/32A & 95.7 & 76.15 \\
    ResNet-DeepLab-TV (ours) & 32W/32A & 95.8 & 76.29 \\
    ResNet-DeepLab     & 5W/5A &  95.6 & 75.92 \\
    ResNet-DeepLab-TV (ours) & 5W/5A &  95.7 & 76.09 \\
    \bottomrule
\end{tabular}
    \caption{Results of image semantic segmentation on Cityscapes using our TV layers. mIoU stands for Mean Intersection over Union.}
    \label{tab:totalVariationSegmentation}
\end{table}

We further demonstrate the contribution of our TV layers on the semantic segmentation task, as it is an important task for a variety of low computational power devices such as robots, drones, and automobiles. To this end, we adopt the popular segmentation architecture DeepLabV3~\cite{chen2017atrous}. Specifically, we construct a DeepLabV3 architecture with a ResNet50~\cite{he2016deep} encoder and the standard ASPP~\cite{chen2017atrous} module as the decoder. 
We train and test our architecture on the popular Cityscapes data-set~\cite{cordts2016cityscapes}, starting with publicly available pre-trained weights\footnote{\url{https://github.com/VainF/DeepLabV3Plus-Pytorch}}. The pre-trained weights were obtained using the standard train and validation split, using SGD optimizer with a batch size of 6, weight decay of 1e-4, and a learning rates of 0.01 for the encoder parameters and 0.1 for the decoder parameters, with an adaptive polynomial learning rate reduction policy. The post training was split to two phases. First, we used the same scheme for the refinement optimization of the quantized net for 30 epochs. Then, after adding the TV layers, we changed the initial learning rate to 0.001 and the batch size to 4 and trained for 10 more epochs. In both phases, the cross-entropy loss is used.
We measure both the accuracy and mean intersection over union (mIoU) metrics, which is defined as:
\[
    \frac{1}{m} \sum_{i=0}^m{\frac{\vert BB_{GT} \cup BB_{pred} \vert}
                                    {\vert BB_{GT} \cap BB_{pred} \vert}}
\]
where $BB_{GT}$ is the bounding box of the ground truth data, $BB_{pred}$ is the bounding box of the prediction and $m$ is the number of samples in a mini-batch. These results are summarized in Tab. \ref{tab:totalVariationSegmentation}. We see that whether the network is quantized or not, incorporating the TV layers to an existing architecture leads to favorable performance.

\subsection{\textbf{Image Classification Using Stable and Quantized Networks}}
\label{sub:experiment_cnns}
In this section, we evaluate our symmetric network architectures under quantization.\\Specifically, we test the theoretical approach presented so far using the ResNet34, ResNet56 \cite{he2016deep} and MobileNetV2~\cite{sandler2018mobilenetv2} architectures, which are popular for image classification tasks and often serve as a benchmark. Their symmetric counterparts, as developed in section \ref{sec:stable_nets}, will serve here to illustrate the theory in practice.  We conduct several experiments using CIFAR-10/100. The networks were trained using SGD for 400 epochs with an initial learning rate of 0.1 and a cosine decay schedule for the learning rate. Quantization was introduced gradually during the training process by starting with 16 bits for both the weights and activations and reducing the number of bits by 1 every ten epochs, until the desired number of bits was reached. Accuracy is computed on the CIFAR-10/100 test data.

\begin{table}
    \centering
    \resizebox{\columnwidth}{!}{
\begin{tabular}{l c c c c c c}
    \toprule
    \multirow{2}{*}{\bfseries Architecture} &
    {\bfseries Params} & 
    \multicolumn{3}{c}{\bfseries Accuracy ($\%$)} & 
    \multicolumn{1}{c}{\bfseries MSE (Acc.)} \\
    & (M) & FP/FP & 4W/8A & 4W/4A & 4A $\rightarrow$ 32A \\ 
    \midrule
    ResNet56 (orig.) & 0.85 & 93.8 & 93.0 & 92.6 & 0.076 (92.9) \\
    Stable ResNet56 (ours) & 0.41 & 92.1 & 92.3 & 91.6 & 0.024 (92.0) \\
    \midrule
    MobileNetV2 (orig.) & 2.20 & 94.0  & 92.9 & 92.4 & 0.067 (92.5) \\
    Stable MobileNetV2 (ours) & 1.70 & 93.1 & 92.2 & 91.6 & 0.034 (91.7) \\
    \bottomrule
\end{tabular}
}
    \caption{Performance on CIFAR-10 image classification. Our stable networks are compared to the original architectures under different quantization levels and their consistency with their non-quantized counterparts is shown in the MSE column. Param. denotes the total number of parameters in each network.}
    \label{tab:stableClassificationCIFAR10}
\end{table}

\begin{table}
    \centering
    \resizebox{\columnwidth}{!}{
\begin{tabular}{l c c c c c c}
    \toprule
    \multirow{2}{*}{\bfseries Architecture} & 
    {\bfseries Params }&
    \multicolumn{3}{c}{\bfseries Accuracy ($\%$)} & 
    \multicolumn{1}{c}{\bfseries MSE (Acc.)} \\
    &{ (M)} & FP/FP & 4W/8A & 4W/4A & 4A $\rightarrow$ 32A  \\ 
    \midrule
    ResNet34 (orig.) & 21.3 & 78.5 & 75.1 & 74.7 & 0.019 (74.7) \\
    Stable ResNet34 (ours) & 9.60 & 76.8 & 75.4 & 75.0 & 0.0083 (75.2) \\
    \midrule
    ResNet56 (orig.) & 0.86 & 72.0 & 69.0 & 69.6 & 0.14 (70.0) \\
    Stable ResNet56 (ours) & 0.41 & 68.4 & 66.7 & 66.1 & 0.031 (67.3) \\
    \midrule
    MobileNetV2 (orig.)  & 2.30 & 74.2 & 71.1 & 69.8 & 0.083 (70.6) \\
    Stable MobileNetV2 (ours) & 1.80 &  73.1 &  71.6 & 70.9 & 0.058 (71.0) \\
    \bottomrule
\end{tabular}
}
    \caption{Performance on CIFAR-100 image classification. Our stable networks are compared to the original architectures under different quantization levels. Their consistency with their non-quantized counterparts is shown in the MSE column. Param. denotes the total number of parameters in each network.}
    \label{tab:stableClassificationCIFAR100}
\end{table}

The results of our experiments, presented in Tables  \ref{tab:stableClassificationCIFAR10} and \ref{tab:stableClassificationCIFAR100}, show that symmetric networks achieve similar accuracy as non-symmetric networks, while using approximately half of the parameters (this number is less than half due to the removal of the $1\times1$ convolutions in the channel-changing blocks as described in section \ref{sub:stable_connector}). This property is important since one of the main goals of network compression is to have low-storage network models. Here, we get a significant saving as a by-product of promoting stability, which is desired for different reasons.

\begin{figure}
\centering
  \begin{subfigure}{0.7\linewidth}
    \centering
    \begin{tikzpicture}
      \begin{axis}[
          width=\linewidth, 
          height=0.6\linewidth,
          grid=major,
          grid style={dashed,gray!30},
          xlabel=Layer,
          ylabel=MSE,
          ylabel near ticks,
          legend style={at={(0.6,-0.25)},anchor=north,scale=0.8, draw=none, cells={anchor=west}},
          legend columns=-1,
          xtick={1,7,14,21,26},
          yticklabel style={
            /pgf/number format/fixed,
            /pgf/number format/precision=3
          },
          scaled y ticks=false,
          every axis plot post/.style={thick},
        ]
        \addplot
        table[x=layer,y=unstable,col sep=comma] {data/resnet56_vs_stablenet56_layerwise_mse_cifar10.csv};

        \addplot
        table[x=layer,y=stable,col sep=comma] {data/resnet56_vs_stablenet56_layerwise_mse_cifar10.csv};
        
        \legend{ResNet56, Sym. ResNet56}
        \end{axis}
    \end{tikzpicture}
    \vspace{1em}
    \end{subfigure}
    \begin{subfigure}{0.7\linewidth}
        \centering
        \begin{tikzpicture}
          \begin{axis}[
              width=\linewidth,
              height=0.6\linewidth,
              grid=major,
              grid style={dashed,gray!30},
              xlabel=Layer,
              ylabel=MSE,
              ylabel near ticks,
              legend style={at={(0.5,-0.25)},anchor=north,scale=0.8, draw=none, cells={anchor=west}},
              legend columns=-1,
              xtick={1,7,14,21,26},
              yticklabel style={
                    /pgf/number format/fixed,
                    /pgf/number format/precision=2
              },
              scaled y ticks=false,
              every axis plot post/.style={thick},
            ]
            
            \addplot
            table[x=layer,y=unstable,col sep=comma] {data/resnet56_vs_stablenet56_layerwise_mse_cifar100.csv};
            
            \addplot
            table[x=layer,y=stable,col sep=comma] {data/resnet56_vs_stablenet56_layerwise_mse_cifar100.csv};
            
            \legend{ResNet56, Sym. ResNet56}
          \end{axis}
        \end{tikzpicture}
        \vspace{0.9em} 
    \end{subfigure}
    
    \begin{subfigure}{0.7\linewidth}
        \centering
        \begin{tikzpicture}
          \begin{axis}[
              width=\linewidth,
              height=0.6\linewidth,
              grid=major,
              grid style={dashed,gray!30},
              xlabel=Layer,
              ylabel=MSE,
              ylabel near ticks,
              legend style={at={(0.5,-0.25)},anchor=north,scale=0.8, draw=none, cells={anchor=west}},
              legend columns=-1,
              xtick={1,6,12,16},
              yticklabel style={
                    /pgf/number format/fixed,
                    /pgf/number format/precision=2
              },
              scaled y ticks=false,
              every axis plot post/.style={thick},
            ]
            \addplot
            table[x=layer,y=unstable,col sep=comma] {data/mobilenetv2_layerwise_mse_cifar100.csv};
            
            \addplot
            table[x=layer,y=stable,col sep=comma] {data/mobilenetv2_layerwise_mse_cifar100.csv};
            \legend{MobileNetV2, Sym. MobileNetV2}
          \end{axis}
        \end{tikzpicture}
    \end{subfigure}
\caption{Per-layer MSE between activation maps of symmetric and non-symmetric network pairs. Each line represents a pair of networks where one has quantized activation maps and the other does not. The values are normalized per-layer to account for the different dimensions of each layer. In all cases, the symmetric variants (in red) exhibit a bounded divergence from full-precision activations, while the non-symmetric networks diverge as the information propagates through the layers (in blue). Hence, they are unstable. Top to bottom: ResNet56/CIFAR-10, ResNet56/CIFAR-100 and MobileNetV2/CIFAR-100. Both networks in each pair achieve comparable classification accuracy.}
\label{fig:stabilityMSE}
\end{figure}
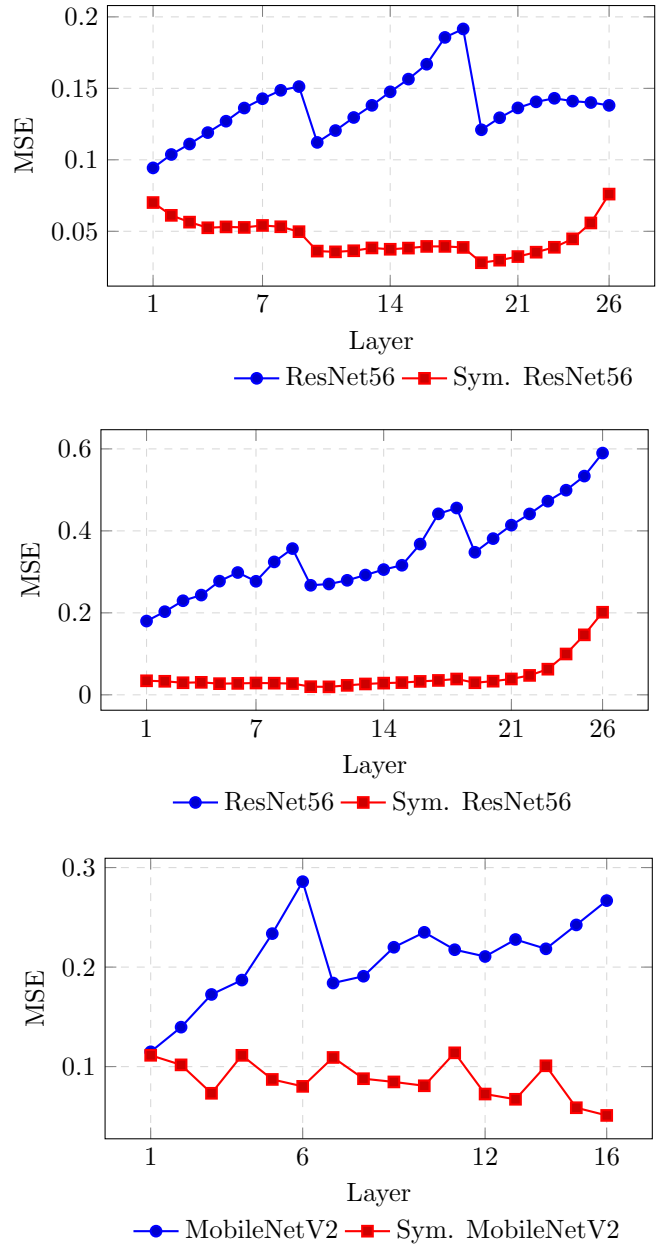

Furthermore, we quantify the notion of stability by comparing the behaviour of symmetric and non-symmetric networks under 4-bit quantization. In addition to the 4-bit quantized network, we relax the quantization of the activation maps, allowing them to use the full 32-bit precision, and we measure the divergence between corresponding activations in both runs of the same network - one at 4 bits and the other at 32 bits. This divergence is summarized as the per-entry MSE between the activation maps throughout the two networks. Tables \ref{tab:stableClassificationCIFAR10} and \ref{tab:stableClassificationCIFAR100} show that the divergence of the two runs is greater in the non-symmetric (original) network variants, as predicted by the theoretical analysis.  We show further evidence in Fig. \ref{fig:stabilityMSE}, where we plot the same MSE difference along the layers of each network. We note an intriguing observation in these plots:  the difference decreases where strided operations are applied. This is not surprising as such down-sampling is equivalent to averaging out quantization errors for which cancellation occurs. Somewhat surprisingly, the networks which were trained using 4-bit activation maps maintain their accuracy when removing the quantization. This is consistent for both the symmetric and non-symmetric networks. Still, having the stable behaviour of the network is a desired property, especially for sensitive applications, as input data in real life may be unpredictable.

\subsection{\textbf{Semi-supervised Node-classification with Quantized GCNs}}
\label{sub:experiment_gcns}
In this section we employ the graph networks PDE-GCN\textsubscript{D}(sym.) and PDE-GCN\textsubscript{D}(non-sym.) from Eq. \eqref{eq:stableDiffusiveGCN}-\eqref{eq:unstableDiffusiveGCN} for semi-supervised node-classification on the Cora, CiteSeer and PubMed data-sets. 
Namely, we measure the impact of symmetry on the network's accuracy, as well as its ability to maintain similar behaviour with respect to activation similarity under activation quantization of 8 and 4 bits. 
In all experiments, we use 32 layers, with 64 hidden channels for Cora, and 256 hidden channels for CiteSeer and PubMed. The training and evaluation procedure we used is identical to~\cite{kipf2016semi}. Details on initialization and hyper-parameter tuning, are given in~\cite{eliasof2021pdegcn}, which we follow in this experiment.
The results provided in Tab. \ref{tab:pdegcn} reveal two benefits of a symmetric (stable) formulation over a non-symmetric (unstable) one. First, it is apparent that that the former results in better accuracy, often by over $2 \%$, while using almost half the number of parameters. In addition, the action of the network is better preserved under quantization using the symmetric formulation. We note that the symmetric formulation in Eq. \eqref{eq:stableDiffusiveGCN} is a sub-set of the non-symmetric counterpart in Eq. \eqref{eq:unstableDiffusiveGCN}. Therefore, theoretically, both networks can achieve identical expressiveness. However, as demonstrated in Tab. \ref{tab:pdegcn}, we did not observe this in practice in any of the experiments we conducted. We attribute this gap to the smoother optimization process of the stable PDE-GCN\textsubscript{D}(sym.).

\begin{table}
    \centering
    \begin{tabular}{lccccc}
    Benchmark & Classes & Label rate & Nodes & Edges & Features \\
    \toprule
    Cora & 7 &  0.052 & 2708 & 5429 & 1433 \\
    CiteSeer & 6 & 0.036 & 3327 & 4732 & 3703 \\
    PubMed & 3 & 0.003 & 19717 & 44338 & 500\\
    \bottomrule
    \end{tabular}
    \caption{Statistics of semi-supervised benchmarks}
    \label{tab:semisupervised_statistics}
\end{table}

\begin{table}
    \centering
    \resizebox{\columnwidth}{!}{
\begin{tabular}{l c c c c c c c c c}
    \toprule
    \multirow{2}{*}{\bfseries Data-set} & 
    \multirow{2}{*}{\bfseries Architecture} & 
    \multirow{1}{*}{\bfseries Params} &
    \multicolumn{3}{c}{\bfseries Accuracy ($\%$)} & 
    \multicolumn{1}{c}{\bfseries MSE}&
    
    \\ 
    & & (M) & FP & 4W/8A & 4W/4A & 4A $\rightarrow$ 32A \\ 
    \midrule
    Cora & PDE-GCN\textsubscript{D} (non-sym.) & 0.35 & 82.7 & 82.2 & 75.7 & 6.11 \\
    &  PDE-GCN\textsubscript{D} (sym.) & 0.22 & 84.3  & 84.0 & 79.4 & 2.03  \\
    \midrule 
    
    CiteSeer & PDE-GCN\textsubscript{D} (non-sym.) & 5.14 & 73.9 & 72.6 & 71.1 & 20.48  \\
    &  PDE-GCN\textsubscript{D} (sym.) & 3.04 & 75.6  & 74.1 &  72.2 &  12.44  \\
    \midrule
    PubMed & PDE-GCN\textsubscript{D} (non-sym.)  & 4.32 & 79.0 & 79.3 & 75.1 & 5.14 \\
    &  PDE-GCN\textsubscript{D} (sym.) & 2.22 & 80.2  & 80.1 & 77.6 & 2.52  \\
    \bottomrule
\end{tabular}
}
    \caption{Comparison of our symmetric PDE-GCN\textsubscript{D} vs the non-symmetric original on node-classification tasks for various data-sets. MSE denotes the similarity of the activations of quantized and full-precision networks. Param. denotes the total number of parameters in each network architecture.}
    \label{tab:pdegcn}
\end{table}

\section{Conclusion}
\label{sec:conclusion}
In this work, we explored quantized neural networks from the perspective of PDE-based concepts and analysis tools, to gain deeper understanding on the nature of quantized CNNs and GCNs. First, we exhibit that using applying TV to the activations yields favorable results, as it promotes smoothness of the activations generated by the network. This is notably beneficial for tasks like semantic segmentation. Then, through a series of experiments, ranging from image classification to graph node-classification, we demonstrate that the concept of stability preserves the action of a network under different quantization rates. In addition, we find that at times, stability aids to improve accuracy. These properties are of particular interest for resource-constrained, low-power or real-time applications like autonomous driving.

\bibliographystyle{spmpsci}
\bibliography{QuantPDE}

\end{document}